\documentclass[runningheads]{llncs}

\usepackage[review,year=2024,ID=*****]{eccv}

\usepackage{eccvabbrv}

\usepackage{graphicx}
\usepackage{booktabs}

\usepackage[accsupp]{axessibility}  

\usepackage[pagebackref,breaklinks,colorlinks,citecolor=eccvblue]{hyperref}

\usepackage{orcidlink}
\usepackage[misc]{ifsym}

\begin{document}
\nolinenumbers
\title{Compound Expression Recognition via Multi Model Ensemble for the ABAW7 Challenge}

\titlerunning{Abbreviated paper title}

\author{Xuxiong Liu\inst{1} \and Kang Shen\inst{1}
Jun Yao \and Boyan Wang  \and Liuwei An \and Zishun Cui \and Minrui Liu \and Xiao Sun\Letter \and
Weijie Feng\Letter}


\institute{Hefei University of Technology\\
Hefei, China\\
\email{\{liuxuxiong, shenkang, yaojun, wangboyan\}@mail.hfut.edu.cn}\\
\email{\{2022171285,anliuwei,liumr\}@mail.hfut.edu.cn}\\
\email{sunx@hfut.edu.cn,wjfeng@hfut.edu.cn}}

\maketitle

\begin{abstract}
Compound Expression Recognition (CER) is vital for effective interpersonal interactions. Human emotional expressions are inherently complex due to the presence of compound expressions, requiring the consideration of both local and global facial cues for accurate judgment. In this paper, we propose an ensemble learning-based solution to address this complexity. Our approach involves training three distinct expression classification models using convolutional networks, Vision Transformers, and multiscale local attention networks. By employing late fusion for model ensemble, we combine the outputs of these models to predict the final results. Our method demonstrates high accuracy on the RAF-DB datasets and is capable of recognizing expressions in certain portions of the C-EXPR-DB through zero-shot learning.
\end{abstract}

\section{Introduction}
\label{sec:intro}
Facial Expression Recognition (FER) holds a significant position in the field of Artificial Intelligence, as it enables computers to better convey human emotional information, supplementing the crucial role of voice in emotional communication in real life. However, traditional facial expression recognition techniques are typically limited to classifying six basic facial expressions, namely anger, happiness, sadness, surprise, disgust, and fear. In reality, human emotional expressions are far more complex than these predefined categories.To address this challenge, Compound Expression Recognition (CER) has emerged as a part of affective computing. CER is an emerging task in intelligent human-computer interaction and multimodal user interfaces. It requires the automatic recognition of individuals' compound emotional states, which may include combinations of two or more basic emotions, such as fearfully surprised, happily surprised, sadly surprised.

The utilization of multimodal features, including visual, audio, and text features, has been extensively employed in previous ABAW competitions \cite{zafeiriou2017aff,kollias2019face,kollias2019expression,kollias2020analysing,kollias2021affect,kollias2021analysing,kollias2021distribution,kollias2022abaw,kollias2023abaw,kollias2023abaw2,kollias20246th,kollias2023multi,kollias2019deep,kollias20247th}. We can improve the performance in affective behavior analysis tasks by extracting and analyzing these multimodal features.
Specifically, our classification task models can typically be divided into two types: ResNet and Transformers. ResNet is one of the commonly used backbone networks in CNNs. It calculates high-level features of images by sliding convolutional kernels over them, focusing on local features. The innovation of ResNet lies in the introduction of residual connections, which make the training of deep networks more efficient and mitigate the vanishing gradient problem. On the other hand, the Vision Transformer is the first widely applied backbone in the Transformer family. It segments the image into patches and then flattens them into a sequence. By incorporating positional encoding, the Vision Transformer embeds the positional information of each patch into the sequence. Through the Transformer's encoder module, the Vision Transformer can model the positional relationships of all locations in the image simultaneously, capturing contextual information from different parts of the face and obtaining global information. \cite{wang2022emotional,HaifengChen2021TransformerEW,wang2022multi} indicate that hybrid models combining CNN and Transformer architectures also demonstrate significant potential. These models can maintain efficient feature extraction while further enhancing the ability to capture global information, thereby achieving superior performance in tasks such as facial expression recognition.

In this paper, we adopt a multi-model solution to address the problem of compound expression recognition. First, we use ResNet50 as the convolutional neural network model, which focuses on capturing the local features of facial expressions. Simultaneously, we employ the ViT to extract features from images, effectively capturing the global information of facial expressions through the self-attention mechanism. Subsequently, we use a multilayer perceptron (MLP) to fuse the features extracted by both models, leveraging the complementarity of local and global features. Finally, we train and validate our model on two commonly used facial expression recognition datasets, RAF-DB and C-EXPR-DB. Through this multi-model fusion approach, we aim to improve the accuracy and robustness of compound expression recognition.

\section{Related Work}
\subsection{Facial Expression Recognition}
Facial Expression Recognition (FER) has made significant progress in its development from recognizing single expressions to compound expressions recognition (CER). CER has garnered widespread attention due to its ability to identify complex facial expressions that convey combinations of basic emotions, reflecting more nuanced human emotional states\cite{dong2024bi,he2022compound}. Research has shown that recognizing basic emotional expressions through deep learning methods paves the way for more advanced approaches capable of deciphering compound expressions\cite{li2019separate,slimani2019compound,zhao2019convolutional,kollias2023multi}.

Typical approaches focus on utilizing Convolutional Neural Networks (CNNs) for feature extraction and employing Recurrent Neural Networks (RNNs) or attention mechanisms to capture the subtle nuances and dynamics of facial expressions over time. Multi-task learning frameworks have also been widely explored to simultaneously recognize multiple basic expressions more accurately and robustly. Researchers have also developed the Real-world Affective Faces (RAF-CE) database, which contains a rich set of compound expression samples. The combination of Meta Multi-task Learning (MML) and Action Units (AU) recognition has significantly improved the performance of compound FER. This approach enhances the model's generalization ability by simultaneously learning multiple related tasks, thereby better understanding the subtle nuances in complex emotions. Additionally, the C-EXPR-DB dataset and the CEXPR-NET model have improved the performance of compound emotion recognition through a multi-task learning approach, utilizing cross-entropy and KL divergence to update the model. Deep Bi Manifold CNN (DBMCNN) introduces a novel manifold-based deep learning network for learning and recognizing compound expressions. Recent advancements include the application of transformer models in emotion recognition, such as Former-DFER, Spatio-Temporal Transformer (STT), and NR-DFERNet, which have improved the performance of dynamic facial expression recognition by capturing both spatial and temporal features. Despite these breakthroughs in addressing discrete label dynamic facial expression recognition (DFER), interference from image backgrounds remains a challenge. To address this issue, researchers have incorporated ensemble learning into their methods to further enhance recognition accuracy and robustness.

\section{Feature Extraction}
We fuse features from different neural networks to obtain more reliable emotional features and utilize these fused features for downstream tasks. By combining information from various feature extraction models such as ResNet and POSTER, we achieve a more comprehensive and accurate representation of emotions.
\subsection{Resnet-18}
ResNet\cite{he2016deep}(He et al. 2016) is a deep convolutional neural network (CNN) architecture designed to address the common issues of vanishing gradients and exploding gradients during the training of deep neural networks. Its core idea is the introduction of residual blocks, which incorporate skip connections, making the network easier to train. Instead of directly learning the mapping of each layer, the output of the residual block learns the residual between the input and output. This structure effectively mitigates the vanishing gradient problem. The pre-trained model of ResNet-18 can be used as a feature extractor, which first pretained on the MS-Celeb-1M\cite{guo2016ms}, and finally obtain a 512-dimensional visual feature vector. transforming images into high-dimensional feature vectors for use in other machine learning tasks, such as image retrieval and image similarity computation.
\subsection{POSTER}
The two-stream Pyramid crOss-fuSion TransformER network (POSTER)\cite{zheng2022poster,mao2023poster} is a novel deep learning model specifically designed for video understanding tasks, such as action recognition and video classification. POSTER combines a pyramid structure with a two-stream architecture, leveraging cross-layer fusion and transformer networks to enhance video understanding performance. Extensive experimental results demonstrate that POSTER outperforms SOTA methods on RAF-DB with 92.05\%, AffectNet\cite{mollahosseini2017affectnet} (7 cls) with 67.31\%, and AffectNet (8cls) with 63.34\%, respectively . The dimension of the visual feature vectors is 768.

\begin{figure}[tb]
  \centering
  \includegraphics[width=12cm]{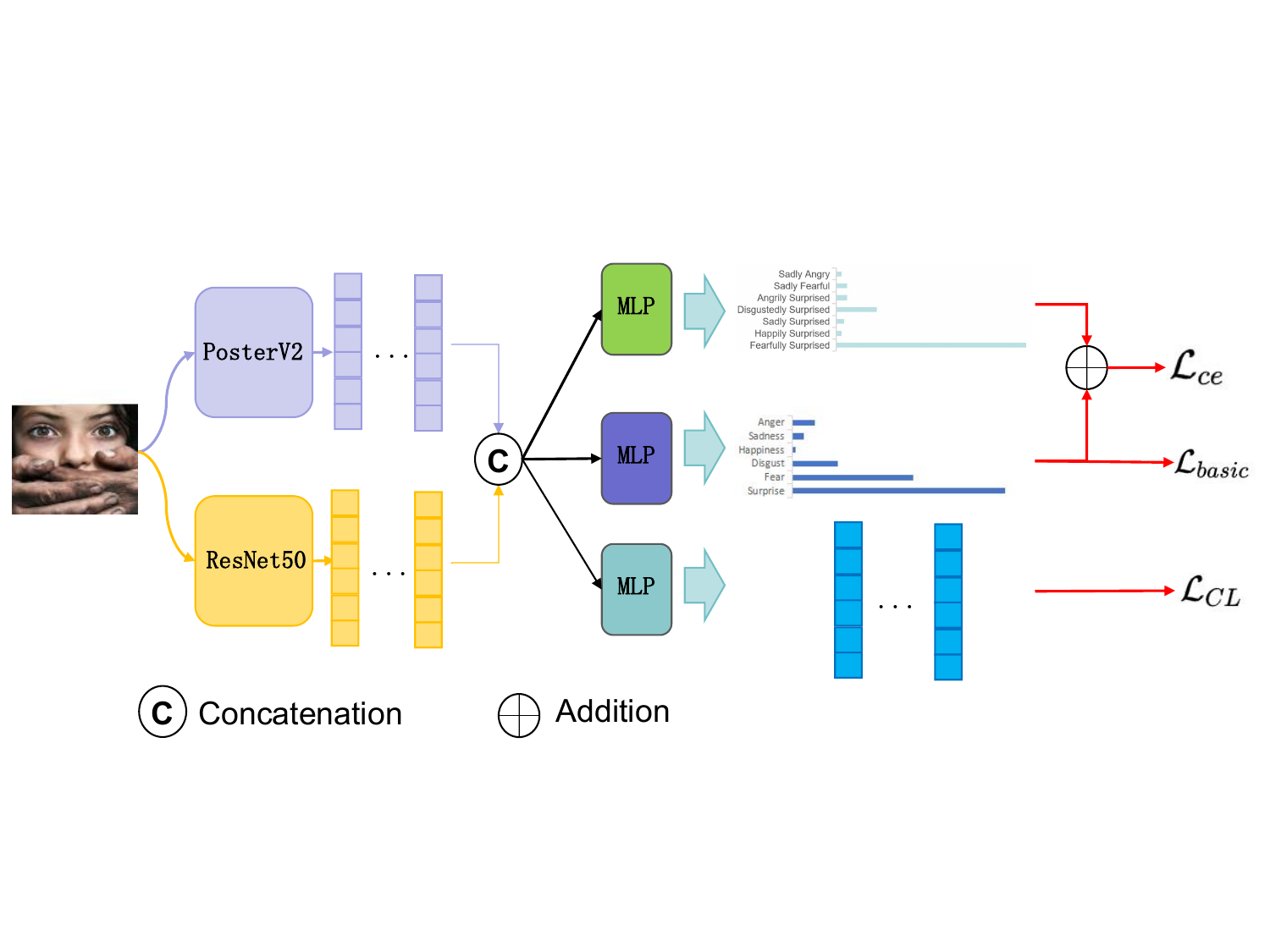}
  \caption{The image and its two distinct enhanced versions are processed through the PosterV2 model and the ResNet50 model to obtain 768-dimensional and 2048-dimensional features, respectively. These features are then concatenated as input for the subsequent stage. The features extracted from the original image are separately input into two Multi-Layer Perceptron (MLP) to obtain basic and compound expression predictions. The basic expression predictions are learned through the loss function $\mathcal{L}_{basic}$. Concurrently, the basic expression predictions are combined with the compound expression predictions and learned through another loss function $\mathcal{L}_{ce}$. Additionally, to better adapt the model to various datasets, we employ contrastive learning as an optimization technique by the loss function $ \mathcal{L}_{CL}$.
  }
  \label{fig:model}
\end{figure}

\section{Method}
In this section, we introduce our late-fusion ensemble model developed for compound expression recognition. As illustrated in  \cref{fig:model}, the pipeline includes two separate models: Vision Transformer (ViT) and ResNet.
\subsection{Data collection}
Given the differences between the ImageNet dataset and facial expression recognition datasets, as well as the limited number of datasets with compound expression annotations like RAF-DB, we built a unified dataset using single-expression annotations from AffectNet and RAF-DB . This dataset comprises a total of 306,989 facial images, with 299,922 used for training and 7,067 for validation.
\subsection{Encoders}
\subsubsection{PosterV2}
We utilized a pre-trained PosterV2, which was initially trained on the ImageNet-1K dataset using self-supervised learning with masked reconstruction. Subsequent experiments in the paper have demonstrated its effectiveness across various image-based tasks, achieving comparable or even superior generalization performance compared to supervised training, including in the domain of emotion recognition for facial expressions. The model processes the extracted facial images and produces 768-dimensional embeddings for each image.
\subsubsection{Residual Neural Network (ResNet)}
As a member of the residual network series, ResNet is widely used in the field of computer vision, including tasks such as image classification, object detection, and image segmentation. Due to its powerful feature extraction capabilities and relatively moderate computational complexity, ResNet50 has become one of the classic backbone networks. We use the weights trained on FER2013 as initialization parameters, generating a 2048-dimensional vector for each image.
\subsection{Ensemble}
We use batched data images x as input to the model, where $X\in \mathbb{R} ^{B\times 3\times H\times W} $.  Here, B denotes the batch size, 3 represents the RGB channels, and H and W are the height and width of the images, respectively. Therefore, the features after data augmentation can be represented as:
\begin{equation}
    feature_1=PosterV2(x)\in  \mathbb{R} ^{B\times 768} 
\end{equation}
\begin{equation}
    feature_2=Resnet(x)\in  \mathbb{R} ^{B\times 512} 
\end{equation}

To improve model performance by merging multiple feature maps for more comprehensive feature representation, we utilize a late fusion strategy. Specifically, we concatenate the three aforementioned features along a designated dimension, then feed these feature maps into a multi-layer perceptron (MLP) and apply softmax to calculate the logits for seven compound expressions, as follows:
\begin{equation}
    feature=\left [ feature_{1} ; feature_{2} \right ] 
\end{equation}
where $[ ; ; ]$ denotes the concatenation operation.
\begin{equation}
    logit=softmax(MLP(feature))
\end{equation}

\section{Experiments}
\subsection{Dataset}
C-EXPR-DB\cite{kollias2023multi} is currently the largest and most diverse audiovisual database collected in the wild. It contains 400 videos, amounting to around 200,000 frames, meticulously annotated for 12 compound expressions and various affective states. Additionally, C-EXPR-DB provides annotations for continuous valence-arousal dimensions, speech detection, facial landmarks, bounding boxes, 17 action units, and facial attributes. In the Compound Expression Recognition Challenge, a total of 56 unlabeled videos were selected, covering 7 types of compound expressions. The extracted video tags include seven compound expressions: Fearfully Surprised, Happily Surprised, Sadly Surprised, Disgustedly Surprised, Angrily Surprised, Sadly Fearful, and Sadly Angry.

We utilize the RAF-DB\cite{li2017reliable} dataset for pretraining our visual feature extractors. The RAF-DB is a large-scale database comprising approximately 30,000 facial images and 3954 compound expression from thousands of individuals. Each image has been independently annotated around 40 times and then filtered using the EM algorithm to remove unreliable annotations.  About half of these images have been manually annotated for seven discrete facial expressions as well as the intensity of valence and arousal in the facial expressions.

\subsection{Implement Details}
\subsubsection{Evaluation metric}
For the Compound Expression Recognition Challenge, the evaluation metric is the F1 Score for seven compound expressions. This metric measures the prediction accuracy of the model for each expression category by combining both precision and recall. It provides a more comprehensive assessment of the model’s overall performance. The metric can be defined as follows:
\begin{equation}
F_1=\sum_{i=1}^7\frac{F_1^i}{7}
\end{equation}
where \( F_1^i \) corresponds to the \( i\)-th expression.

\subsubsection{Training setting}
All experiments in this paper are conducted using PyTorch and trained on a 64-bit Linux computer with a 64 AMD Ryzen Threadripper 3970X 32-core CPU (3.80GHz), 64 GB RAM, and a 24GB NVIDIA RTX 3090 GPU. The input image resolution is consistently set to 224 × 224 pixels. The training process runs for 100 epochs, utilizing cross-entropy loss as the optimization objective and the Adam optimizer for parameter updates. To improve stability during training, a warm-up learning rate strategy is employed for pre-training the ViT on Unity. In the RAF-DB compound expression experiment, the learning rate is set to 5e-5, and the batch size is set to 128.

\subsection{Results}

\subsubsection{Visual Models’ performance on RAF-DB} \cref{tab2} presents a performance comparison of ViT, Poster, and ResNet in recognizing compound expressions on the RAF-DB validation set. For each compound expression, the table lists the recognition accuracy of the three models, along with their overall accuracy and F1 scores. 
In recognizing the Happily Surprised expression, ViT leads with an accuracy of 92.59\%. For the Sadly Surprised expression, ResNet significantly outperforms both ViT and Poster with an accuracy of 55.56\%, compared to 38.89\% for the latter two. 
Overall, ViT achieves the highest accuracy at 78.09\%, followed by ResNet at 75.06\%, and Poster at 74.06\%. Regarding F1 scores, ViT and ResNet perform similarly with scores of 70.25\% and 68.19\%, respectively, while Poster has the lowest performance at 63.57\%.
These results indicate that ViT demonstrates a well-balanced performance overall, though it may not be the best for certain individual expressions. This analysis highlights that different network architectures exhibit unique strengths and weaknesses in recognizing complex emotional expressions. Therefore, it suggests the potential benefit of combining multiple models to enhance the accuracy and robustness of expression recognition in practical applications.

\begin{table}[tb]
  \caption{Font sizes of headings. 
    Table captions should always be positioned \emph{above} the tables.
  }
  \label{tab2}
  \centering
  \begin{tabular}{@{}cccc@{}}
    \toprule
    Compound Expression & ViT & PosterV2& ResNet\\
    \midrule
    Angrily Surprised  & 60.53 &52.63& 55.26\\
    Disgustedly Surprised & 51.43 & 37.14&60\\
    Fearfully Surprised & 80.17& 81.03& 75.86\\
    Happily Surprised & 92.59 & 89.63& 85.93\\
    Sadly Angry & 84.85 &75.75 &84.85\\
    Sadly Fearful &72.72 &63.64 & 63.64\\
    Sadly Surprised & 38.89 & 38.89 & 55.56\\
    \midrule
    acc & 62.13 & 64.34 &60.35 \\
    \midrule
    F1 &  54.23 & 55.36 & 51.54 \\
  \bottomrule
  \end{tabular}
\end{table}

\begin{table}[tb]
  \caption{Font sizes of headings. 
    Table captions should always be positioned \emph{above} the tables.
  }
  \label{tab3}
  \centering
  \begin{tabular}{@{}cc@{}}
    \toprule
    Compound Expression & Ensemble \\
    \midrule
    Angrily Surprised  & 50.00\\
    Disgustedly Surprised & 45.71\\
    Fearfully Surprised & 87.14\\
    Happily Surprised & 85.93\\
    Sadly Angry & 84.84 \\
    Sadly Fearful &77.27\\
    Sadly Surprised & 27.78\\
    \midrule
    acc & 73.80 \\
    \midrule
    F1 & 63.79  \\
  \bottomrule
  \end{tabular}
\end{table}
\subsubsection{Visual Models’ performance on RAF-DB (CE) }
The results of the ensemble models with late fusion on the RAFDB dataset are shown in \cref{tab3}, and it can be seen that Compared with the single model, the integrated model is more accurate in predicting five compound expressions, namely,Angry Surprised, Disgustedly Surprised, Disgustedly Surprised, Sadly Fearful, and Sadly Surprised. In particular, the expression of Sadly Surprised, which is difficult to identify, is 22.22\% higher than that of ViT. In addition, the accuracy rate and F1 score are both better, which is consistent
with our idea of using different models to bridge the gap between each other. 
\section{Conclusion}
In this paper, we present our solution to the 7th Affective Behavior Analysis in the Wild Workshop and Competition (ABAW), focusing on the recognition of complex expressions. We train three models for expression classification: one based on convolutional networks, one on visual transformers, and one on multi-scale local attention networks. By employing model ensemble techniques with late fusion, we combine the outputs of these models to predict the final result. Extensive experiments demonstrate that our method significantly outperforms the baseline and achieves excellent results in the competition.

%
%
\bibliographystyle{splncs04}
\bibliography{main}
\end{document}